\documentclass[lettersize,journal]{IEEEtran}
\usepackage{amsmath,amsfonts}
\usepackage{algorithmic}
\usepackage{algorithm}
\usepackage{array}
\usepackage{subfigure}
\usepackage{textcomp}
\usepackage{stfloats}
\usepackage{url}
\usepackage{verbatim}
\usepackage{graphicx}
\usepackage{cite}
\usepackage{amssymb}
\usepackage{booktabs}
\usepackage{tabularx}
\usepackage{multirow}
\usepackage{hyperref}   % 用于生成超链接
\usepackage{orcidlink} %调包
\hypersetup{hidelinks} % hide the hyperlink box
\hypersetup{
    colorlinks=true,    % 设置为 true 使链接有颜色
    linkcolor=blue,     % 内部链接的颜色，比如目录
    citecolor=blue,     % 引用的颜色
    filecolor=magenta,  % 文件链接颜色
    urlcolor=black      % URL 链接颜色
}
\begin{document}
\begin{sloppypar}

\title{Capture Artifacts via Progressive Disentangling and Purifying Blended Identities for Deepfake Detection}

\author{Weijie Zhou$^{\orcidlink{0009-0008-9758-9224}}$, Xiaoqing Luo$^{\orcidlink{0000-0001-8030-1660}}$,~\IEEEmembership{Member,~IEEE,} Zhancheng Zhang$^{\orcidlink{0000-0002-7729-6896}}$, Jiachen He, \\and Xiaojun Wu$^{\orcidlink{0000-0002-0310-5778}}$,~\IEEEmembership{Member,~IEEE,}
% <-this % stops a space

\thanks{Weijie Zhou, Xiaoqing Luo, Jiachen He, and Xiaojun Wu are with the School of Artificial Intelligence and Computer Science, Jiangnan University, Wuxi 214122, China.

Zhancheng Zhang is with the School of Electronic and Information Engineering, Suzhou University of Science and Technology, Suzhou 215009, China.

Corresponding author: Xiaoqing Luo, xqluo@jiangnan.edu.cn.}% <-this % stops a space
}

\markboth{IEEE TRANSACTIONS ON CIRCUITS AND SYSTEMS FOR VIDEO TECHNOLOGY}%
{Shell \MakeLowercase{\textit{et al.}}: A Sample Article Using IEEEtran.cls for IEEE Journals}


\maketitle

\begin{abstract}
The Deepfake technology has raised serious concerns regarding privacy breaches and trust issues. To tackle these challenges, Deepfake detection technology has emerged. Current methods over-rely on the global feature space, which contains redundant information independent of the artifacts. As a result, existing Deepfake detection techniques suffer performance degradation when encountering unknown datasets. To reduce information redundancy, the current methods use disentanglement techniques to roughly separate the fake faces into artifacts and content information. However, these methods lack a solid disentanglement foundation and cannot guarantee the reliability of their disentangling process. To address these issues, a Deepfake detection method based on progressive disentangling and purifying blended identities is innovatively proposed in this paper. Based on the artifact generation mechanism, the coarse- and fine-grained strategies are combined to ensure the reliability of the disentanglement method. Our method aims to more accurately capture and separate artifact features in fake faces. Specifically, we first perform the coarse-grained disentangling on fake faces to obtain a pair of blended identities that require no additional annotation to distinguish between source face and target face. Then, the artifact features from each identity are separated to achieve fine-grained disentanglement. To obtain pure identity information and artifacts, an Identity-Artifact Correlation Compression module (IACC) is designed based on the information bottleneck theory, effectively reducing the potential correlation between identity information and artifacts. Additionally, an Identity-Artifact Separation Contrast Loss is designed to enhance the independence of artifact features post-disentangling. Finally, the classifier only focuses on pure artifact features to achieve a generalized Deepfake detector. Extensive experimental evaluations show that the proposed method achieves superior detection performance and better generalization compared to existing state-of-the-art methods.
\end{abstract}


% figure1
\begin{figure}[t]
    \centering
    \includegraphics[width=1\linewidth]{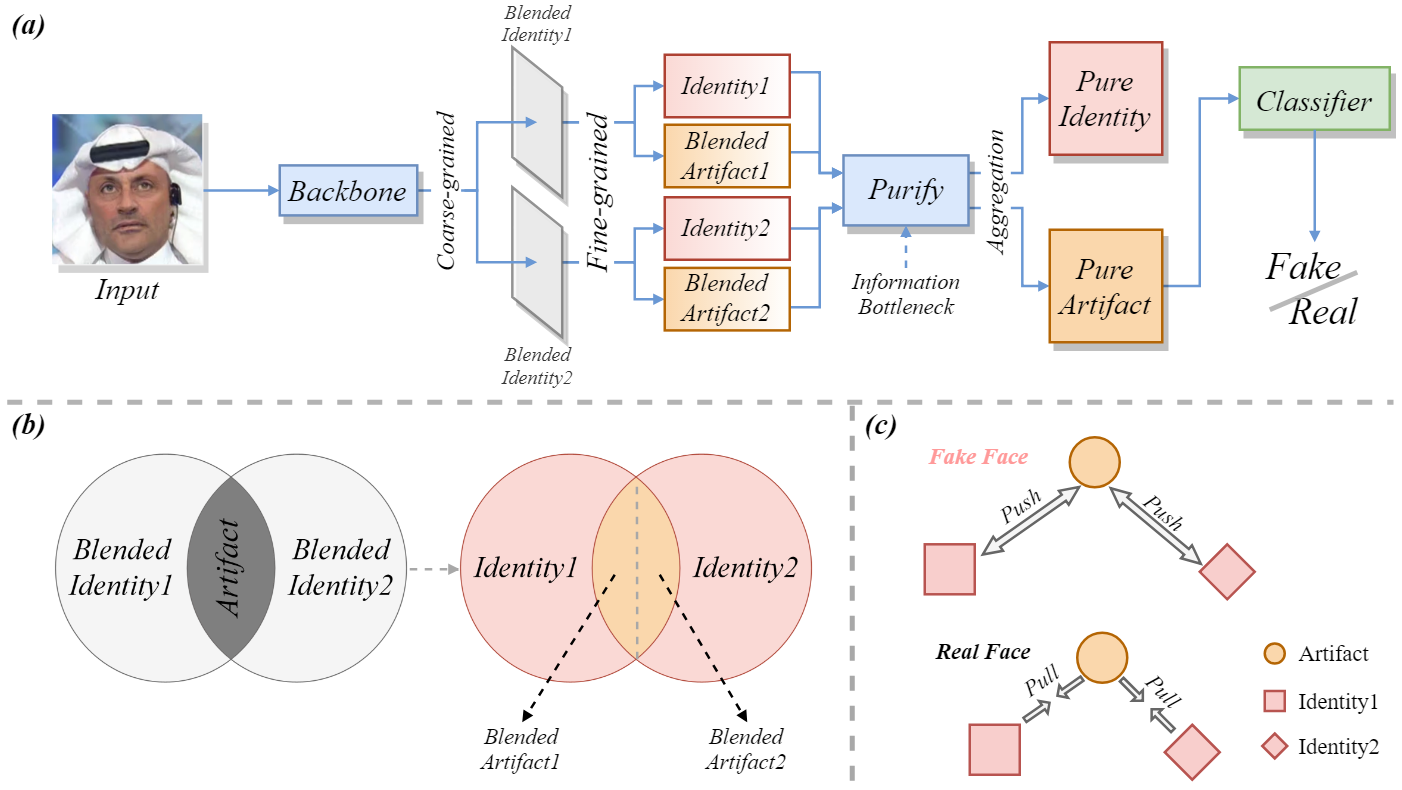}
    \caption{(a) The flowchart of the proposed Deepfake detection method based on progressive disentanglement; (b) Venn diagram illustrating the identity-artifact relationship, supporting the rationale and process of the disentanglement approach; (c) The effect of the Identity-Artifact Separation Contrastive loss: pushing the pure artifact features away from the identity of the fake face, while pulling the identity of the real face closer to the pseudo artifact features.}
    \label{fig:fig_1}
\end{figure}

\section{Introduction}
\IEEEPARstart{D}{eepfake} technology, with its remarkable ability to generate highly realistic visual content, has rapidly garnered significant attention and recognition in both academic research and practical applications \cite{1bitouk2008face, 3korshunova2017fast, 4deepfakes}. However, this technology also presents potential negative effects, as it can be maliciously used to invade personal privacy and spread misinformation, leading to an erosion of public trust in digital media \cite{6lyu2020deepfake}. Given the possible negative impacts, developing an efficient, accurate, and reliable method for detecting fake faces is of urgent importance.

Most previous Deepfake detection techniques \cite{7afchar2018mesonet, 8rossler2019faceforensics++} have shown excellent detection performance on in-domain datasets, but their accuracy drops sharply when faced with unseen datasets or more complex real-world scenarios. Therefore, improving the generalization capability of Deepfake detectors is a challenge faced by researchers. To address this challenge, more and more methods \cite{10li2020face, 11li2018exposing, 13zhao2021multi} focus on the various artifacts generated during the fake face creation process. However, each type of forgery produces specific artifacts, and overfitting to these specific artifacts still severely limits the generalization ability of these methods. Furthermore, these approaches tend to over-rely on the global feature spaces during processing, which are filled with a large amount of redundant information unrelated to the artifacts, such as identity information \cite{14dong2023implicit}, background information \cite{15liang2022exploring}, and so on. This redundant information significantly affects the discriminative power of the detector.

Recently, several effective methods \cite{15liang2022exploring, 16hu2021improving, 56li2022artifacts, 17yan2023ucf} have employed disentanglement techniques to address these issues, forcing the model to focus solely on artifacts while removing the influence of irrelevant information. Therefore, how to achieve disentanglement has become the primary problem to be solved. Existing disentanglement methods, inspired by style transfer \cite{18huang2018multimodal}, crudely divide fake faces into content and artifacts. UCF \cite{17yan2023ucf} argues that previous disentanglement methods overly rely on specific forgery patterns, limiting their generalization ability. As a solution, UCF refines artifacts into specific and common artifacts, using only common artifacts for detection. However, these methods still cannot ensure the reliability of the disentanglement process. Notably, unlike style transfer tasks, where style and content images are independent, the Deepfake detection task only disentangles a single fake face image, making it difficult to guarantee the orthogonality between disentangled components. Directly classifying the disentangled artifacts may interfere with the model judgment. Therefore, the new challenge is to  design a reliable disentanglement method and effectively reduce the correlation between disentangled components to improve the detector accuracy. 

According to the basic principles of face synthesis \cite{3korshunova2017fast}, during the generation of a fake face, two independent face identities become coupled to each other, leading to identity inconsistencies in the fake face \cite{14dong2023implicit}. Therefore, artifacts are defined as information from one identity mixed into another. In this case, a fake face can be defined as a composition of two blended identities, with each blended identity containing blended artifacts coupled with it. Inspired by this observation, we propose a new Deepfake detection method based on progressive disentangling and purifying blended identities (as shown in Fig. \ref{fig:fig_1}(a)). The disentanglement framework fully considers the artifact generation mechanism and combines both coarse- and fine-grained strategies to ensure the reliability of the disentanglement process. Specifically, we first perform coarse-grained disentanglement on the fake face, obtaining a pair of blended identities without requiring additional annotation to distinguish between the source and target identities. These two identities stem from the different faces involved in the face synthesis process. Subsequently, the artifact features are further separated from each blended identity, achieving the goal of fine-grained disentanglement. To obtain pure identities and artifacts, an \textbf{Identity-Artifact Correlation Compression (IACC) module} based on \textbf{Information Bottleneck theory} is designed, which effectively reduces the correlation between identities and artifacts. Additionally, an \textbf{Identity-Artifact Separation Contrastive loss} is introduced to enhance the independence of disentangled artifact features (as shown in Fig. \ref{fig:fig_1}(c)). Finally, the pure identities and pure artifacts are aggregated, transforming the representation of a fake face into a set of pure identity and artifacts. The proposed method ensures that the detector focuses solely on pure artifact features, thus eliminating the impact of identity information on the results.

In summary, the main contributions of this study can be summarized as follows:
% contributions
\begin{itemize}
\item We thoroughly analyze the generation mechanism of artifacts and proposed a novel progressive blended identity disentanglement framework. By combining coarse-grained and fine-grained strategies, the fake faces are decoupled into identity information and artifacts, ensuring the reliability of the disentanglement approach. 
\item To reduce the coupling between identity information and artifacts, an Identity-Artifact Correlation Compression (IACC) module based on the Information Bottleneck theory is designed, thereby obtaining pure artifacts and eliminating the influence of identity information on the discrimination results.
\item The proposed Identity-Artifact Separation Contrastive loss is further proposed to enhance the independence of artifact features after disentanglement. Extensive experimental evaluations are conducted on DeepfakeBench \cite{21yan2023deepfakebench}, and the results clearly demonstrate that the progressive disentanglement method provides substantial improvements over baseline methods and outperforms recent competitive detectors.
% \item Extensive experimental evaluations are conducted on DeepfakeBench \cite{21yan2023deepfakebench}, and the results clearly demonstrate that the progressive disentanglement method provides substantial improvements over baseline methods and outperforms recent competitive detectors.
\end{itemize}

% figure2
\begin{figure*}[t]
    \centering
    \includegraphics[width=1\linewidth]{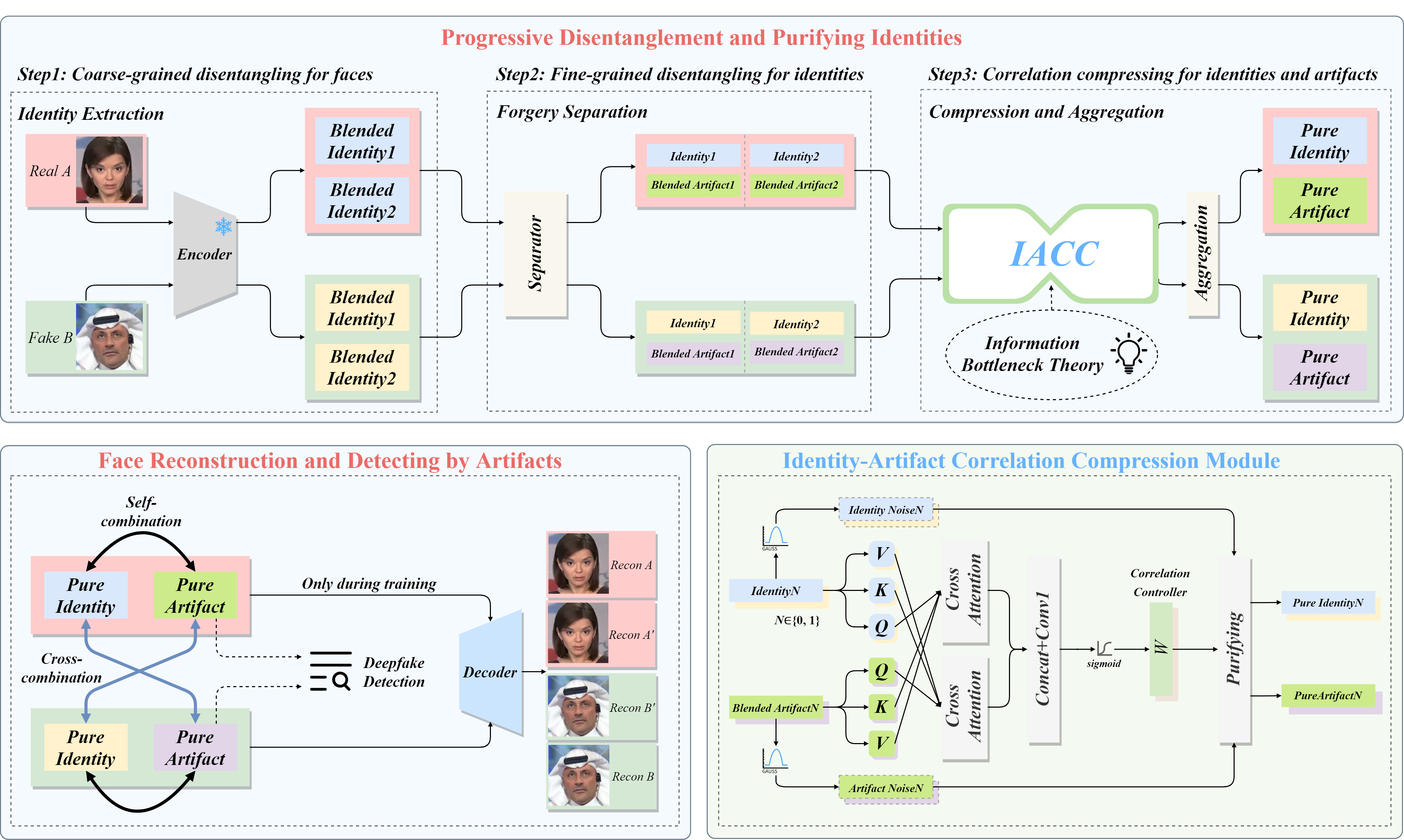}
    \caption{Framework diagram of the Deepfake detection method based on progressive disentangling and purifying blended identities. The network input is a pair of images, consisting of a real face image and a fake face image. The main components of the framework are as follows: 
1) Identity Extraction (Coarse-Grained). 
2) Artifact Separation (Fine-Grained). 
3) Correlation Compression and Aggregation. 
4) Face self-reconstruction and cross-reconstruction. 
5) Deepfake Detection.}
    \label{fig:fig_2}
\end{figure*}

\section{Related Work}
\textbf{General Deepfake Detection.} Most previous Deepfake detection techniques have focused on prominent visual artifacts produced during face forgery, such as blinking \cite{22li2018ictu}, head pose \cite{23yang2019exposing}, lip movement \cite{24haliassos2021lips, 57wang2023exploiting}, and synthetic boundary artifacts \cite{10li2020face, 25shiohara2022detecting}. Beyond these spatial-domain artifacts, FTCN \cite{26zheng2021exploring} combined spatial and temporal domains to explore inconsistencies in facial sequences, such as unnatural mouth movements and discontinuities in facial regions. However, these methods heavily rely on spatial-domain artifacts, making them susceptible to common disturbances like image or video compression and noise. To enhance the robustness of detection, SRM \cite{27luo2021generalizing} leveraged the correlation between RGB features and high-frequency noise features to improve artifact detection. Due to the performance limitations caused by data diversity, some methods use data augmentation to address data insufficiency. Hu et al. \cite{60hu2021detecting} proposed a two-stream network that combines a frame-level stream and a temporality-level stream, where temporal features are extracted through the frame-level stream, while the temporal-level stream captures inconsistency between frames. DCL \cite{28sun2022dual} employed data augmentation and contrastive learning to enable the model to learn richer and more complex representations of forged faces, improving the generalization ability, but the performance still declines significantly on datasets with unknown forgery methods. RECCE \cite{29cao2022end} attempted to teach the model to understand unknown types of forgery by reconstructing real faces. Additionally, Wu et al. \cite{58wu2023interactive} proposed an Interactive Two-Stream Network (ITSNet), where high-frequency cues are extracted using a Decomposable Discrete Cosine Transform (DDCT), and an interaction module is utilized to enable communication across different modalities. Considering that fake faces contain two inconsistent identities, IID \cite{31huang2023implicit} introduced an implicit identity-driven framework for face swapping detection to distinguish explicit and implicit identities for detection. Despite achieving good results on existing datasets, these methods overly depend on the global feature space, causing models to  focus on irrelevant redundant content, severely limiting their generalization ability.Since the forgery traces in the noise domain are complementary to the tampering artifacts present in the image domain, Zhang et al. \cite{59zhang2024face} proposed a face forgery detection network that combines spatial and noise domains. This approach leverages dual-domain fusion and local enhancement to achieve a more comprehensive feature representation.

\textbf{Deepfake Detection Based on Disentanglement Learning. }Disentanglement learning is a method that decomposes coupled dimensional information in complex features into simple, discriminative features and encodes these features into independent dimensions \cite{15liang2022exploring, 32bengio2013representation}. To address the issue of redundant information, existing research has introduced the disentanglement techniques into the Deepfake detection domain, aiming to separate the fake faces into features related to artifacts and irrelevant content, thereby mitigating the impact of the irrelevant content on the detection results. Hu et al. \cite{16hu2021improving} proposed a disentanglement framework that uses only the features related to fake operations for detection. To ensure the independence of disentangled features, Liang et al. \cite{15liang2022exploring} employed content consistency and global representation contrastive constraints. UCF \cite{17yan2023ucf} distinguished between common and specific forgery modes to prevent the model from over-relying on particular fake methods. Although these disentanglement methods, inspired by style transfer tasks, partially address the problem of redundant information, they still fail to fully eliminate the potential correlations between disentangled components. Therefore, developing a disentanglement method specifically designed for Deepfake detection is crucial for enhancing the generalization ability of detection methods.

\section{Methodology}
\subsection{Motivation}
Existing Deepfake detection methods perform well on known fake face datasets but exhibit poor generalization on datasets with unknown forgery modes. The main reason for  this generalization issue is that the current methods excessively rely on the global feature space, which contains a significant amount of redundant information unrelated to artifacts. Some approaches have been proposed using disentanglement techniques \cite{16hu2021improving, 17yan2023ucf} to view the fake faces as a combination of artifacts and content information, forcing the detector to focus solely on the artifact and thus eliminating the effect of content information. However, this style transfer-inspired disentanglement is not fully applicable to Deepfake detection, nor does it guarantee the reliability of the disentanglement approach, resulting in potential correlations between the disentangled components that affect the discriminative ability of the detector.

To overcome these challenges, we propose a progressive blended identity disentanglement framework based on artifact generation mechanisms, combining coarse- and fine-grained disentanglement strategies to ensure the reliability in Deepfake detection. Analyzing fake face synthesis, we define artifacts as information from one identity blended into another. Due to identity inconsistency in fake faces, our framework first disentangles a fake face into a pair of blended identities. Subsequently, the artifact features are separated from the blended identities, and the Identity-Artifact Correlation Compression (IACC) module is used to reduce the degree of information coupling between them. Finally, only the pure artifact features are detected, eliminating the influence of identity information. Additionally, we design an Identity-Artifact Separation Contrastive loss to further compress the potential correlations between identities and artifacts. Our progressive disentanglement framework ensures reliable disentanglement and enhances the generalization ability of Deepfake detectors.

\subsection{Overview of Progressive Disentanglement}
Based on the artifact generation mechanisms, our progressive disentanglement framework is illustrated in Fig. \ref{fig:fig_2}. This framework comprises an encoder, an artifact separator, an Identity-Artifact Correlation Compression (IACC) module, a decoder, and a classifier. Specifically, two identity encoders are used in the encoder to extract a pair of blended identities without requiring additional annotations. The artifact separator is employed to separate the blended artifacts associated with these identities from the extracted features. The IACC module is utilized to reduce the potential correlations between identity information and artifacts, effectively purifying both. The decoder utilizes the separated clean artifacts as conditional information to reconstruct the face image. Authenticity detection is performed by the classifier based solely on the clean artifacts. Notably, the two identity encoders share the same structure but do not share parameters.

\subsection{Coarse- and Fine-grained Disentanglement}
\textbf{Identity Encoding. }The network takes a pair of face images $(I_A, I_B)$ as input, where $I_A$ represents a real image and $I_B$ represents a fake image. We use the pre-trained EfficientNet-B4 \cite{33tan2019efficientnet} as the encoder $E_{id}$ to perform coarse-grained disentanglement of the faces. The encoder $E_{id}$ consists of two independent identity encoders $E_{id1}$ and $E_{id2}$ designed to extract a pair of blended identities without requiring additional annotations, as follows:
\begin{equation}\label{equ1}
    \begin{gathered}
        \widetilde{id^1_x} = E_{id1}(I_x), \widetilde{id^2_x} = E_{id2}(I_x),
    \end{gathered}
\end{equation}
where $x \in \left\{A, B\right\}$ is the label distinguishing between real and fake images, while $\widetilde{id^1_x}$ and $\widetilde{id^2_x}$ represent the blended identity information obtained from encoding each face image through the identity encoders.

\textbf{Artifact Separation. }The artifact separator $S$ is employed to perform fine-grained disentanglement on the blended identity information $\widetilde{id^1_x}$ and $\widetilde{id^2_x}$ to explore and compress the potential correlations between them. Specifically, the blended artifacts associated with each identity are separated by the separator $S$, as follows:
\begin{equation}\label{equ2}
\left[(\overline{id^1_x},\widetilde{art^1_x}), (\overline{id^2_x},\widetilde{art^2_x})\right]=S(\widetilde{id^1_x},\widetilde{id^2_x}),
\end{equation}
where $\overline{id^1_x}$ and $\overline{id^2_x}$ represent the non-pure identity information obtained after separating the blended identities and artifacts, $\widetilde{art^1_x}$ and $\widetilde{art^2_x}$ indicate the blended artifacts corresponding to each identity. The separation process is visually shown in Fig. \ref{fig:fig_1}(b).

\subsection{Correlation Compression and Aggregation}
The identity information and artifacts obtained from the artifact separator are still non-pure. To achieve pure identity information and pure artifacts, we designed the Identity-Artifact Correlation Compression (IACC) module to explore and compress the potential correlations between identities and blended artifacts. Additionally, we utilize Information Bottleneck (IB) theory \cite{20tishby2000information} to guide and optimize the compression process, ensuring its thoroughness while retaining important information relevant to Deepfake detection as much as possible in the artifact features. Finally, pure identity information and pure artifacts are aggregated respectively, transforming the fake faces into a new collection of pure identity information and pure artifacts.

In this section, the fundamental concepts of Information Bottleneck theory is first outlined. Next, the design of IACC module based on this theory is described. Finally, how to use Information Bottleneck theory to optimize the compression process of IACC and retain important information relevant to Deepfake detection.

\textbf{IB theory. }The Information Bottleneck (IB) theory describes the need for neural networks to find the optimal representation $Z$ of the input $X$ during training, i.e., 1) the model should maintain information relevant to the task goal $Y$, and 2) redundant information unrelated to $Y$ should be compressed at the same time \cite{34tishby2015deep}, thus improving the robustness of the model. Therefore, the objective representation of IB theory is as follows:
\begin{equation}\label{equ3}
\max_{z}(I(Z;Y)-\beta I(X;Z)),
\end{equation}
where $\beta$ is the trade-off parameter, $I(Z;Y)$ response optimal indicates the extent to which information relevant to the task goal $Y$ is retained in $Z$, while $I(X;Z)$ indicates the extent to which input $X$ is compressed with respect to redundant information not relevant to the goal $Y$ in $Z$.

\textbf{IACC module. }In order to mine and compress the potential correlations between non-pure identity information and blended artifacts, a cross attention \cite{35huang2019ccnet} guided information compression module based on the compression process of information bottleneck (IB) theory is designed. The structure of the IACC module is shown in Fig. \ref{fig:fig_2}. For fake faces, we first calculate the potential correlations between each non-pure identity and its associated blended artifact using a Dual Cross-Attention Mechanism (DCAM) to obtain the correlation controller $W$, as shown below:
\begin{equation}\label{equ4}
W=sigmoid[DCAM(\overline{id^n}, \widetilde{art^n})]\in [0,1],
\end{equation}
where $W$ has the same size as $\overline{id^n}$ and $\widetilde{art^n}$, and $n \in \left\{0, 1\right\}$ denotes the identity index. We then compress the information in the identity that has potential correlations with the artifacts by adding noise \cite{36smilkov2017smoothgrad, 37karras2020analyzing}. Specifically, we obtain pure identity information by linearly mixing between $\overline{id^n}$ and the Gaussian noise $\varepsilon ^n_{id}$ via the correlation controller $W$. The same purify operation is applied to the blended artifacts. The purification process can be represented as follows:
\begin{equation}\label{equ5}
    \begin{gathered}
    id^n=(1-W)\times\ \overline{id^n}+W\times\ \varepsilon ^n_{id}, \\
    art^n=(1-W)\times\ \widetilde{art^n}+W\times\ \varepsilon ^n_{art},
    \end{gathered}
\end{equation}
where the means and variances of noise $\varepsilon^n_{id}\sim N(\mu_{\overline{id^n}}, \sigma^2_{\overline{id^n}})$ and $\varepsilon ^n_{art}\sim N(\mu_{\widetilde{art^n}}, \sigma^2_{\widetilde{art^n}})$ are the same as those of $\overline{id^n}$ and $\widetilde{art^n}$. To obtain the pure identity feature $id^n$ and pure artifact feature $art^n$, we use the correlation controller $W$ to replace the parts of the identity information that have potential correlations with the artifact features with Gaussian noise. Similarly, the parts of the artifact features related to the identity information are also replaced by Gaussian noise with the same distribution. This approach effectively eliminates the potential correlations between identity information and artifacts, allowing the model to focus more on learning pure identity and artifact features, thereby enhancing the robustness and generalization of model.

\textbf{Guidance from Information Bottleneck Theory.} Based on Information Bottleneck (IB) theory, we design an information optimization loss function to guide the IACC module in compressing the potential correlations between identity information and artifacts while retaining important information relevant to Deepfake detection. By reducing the mutual information $I$ between non-pure identity $\overline{id^n}$ and blended artifacts $\widetilde{art^n}$, the objective of information compression outlined in IB theory is realized. The objective optimization expression is as follows:
\begin{equation}\label{equ6}
\min  {\textstyle \sum_{n=1}^{2}}I(\overline{id^n},\widetilde{art^n}) ,
\end{equation}

Since our identity encoder $E_{id}$ uses a fixed-parameter pre-trained model, the KL-divergence between the features distributions $\overline{id^n}$ and $\widetilde{art^n}$ remains unchanged during scaling. Therefore, our objective optimization expression can be transformed into:
\begin{equation}\label{equ7}
\max  {\textstyle \sum_{n=1}^{2}}KL[p(\overline{id^n}),p(\widetilde{art^n})] ,
\end{equation}
where $p(z)$ represents the probability distribution of non-fully pure identity information and blended artifact features, and $KL(x, y)$ is used to calculate the KL-divergence divergence between these two distributions.

In addition to compressing the mutual information between identity information and artifacts, since the detector ultimately focuses solely on artifact information, we also need to ensure that the pure artifacts $art^n$ retain as much useful information from the blended artifacts $\widetilde{art^n}$ as possible. According to the optimization objectives proposed in Information Bottleneck (IB) theory, we should maximize the retention of information relevant to the task goal \cite{34tishby2015deep}. Specifically expressed as:
\begin{equation}\label{equ8}
\max  {\textstyle \sum_{n=1}^{2}} I(art^n, \widetilde{art^n}) \triangleq \min  {\textstyle \sum_{n=1}^{2}}KL[p(art^n),p(\widetilde{art^n})],
\end{equation}
Combining Eq. \ref{equ7} and Eq. \ref{equ8}, the proposed Information Optimization loss $L_{Info}$ is expressed as follows:
\begin{equation}\label{equ9}
\begin{aligned}
L_{Info}&=exp(-\sum_{n=1}^{2}KL[p(\overline{id^n}),p(\widetilde{art^n})])\\
&+0.5*exp(\sum_{n=1}^{2}KL[p(art^n),p(\widetilde{art^n})],
\end{aligned}
\end{equation}
The Information Optimization loss function $L_{Info}$ allows us to simultaneously compress the potential correlations between identity information and artifacts while preserving crucial information related to Deepfake detection in the artifacts, thereby enhancing the detection accuracy of model.

\textbf{Information Aggregation.} After obtaining two pairs of pure identity information and pure artifacts through the IACC module, the identities and artifacts are aggregated correspondingly. This transforms the faces into a new set that includes complete pure identity information $ID_x$ and pure artifacts $ART_x$. The information aggregation process is represented as follows:
\begin{equation}\label{equ10}
\begin{gathered}
    ID_x=Concat(id^1_x,id^2_x), \\
    ART_x=Concat(art^1_x,art^2_x), \\
    Face_x=(ID_x,ART_x), \\
\end{gathered}
\end{equation}
where $x \in \left\{A, B\right\}$ is the label distinguishing between real face images and fake face images.

\subsection{Face Reconstruction and Deepfake Detection}
\textbf{Face Reconstruction.} The self-reconstruction and cross-reconstruction of faces are achieved by combining both the identity information and artifacts of the same face, as well as those of different faces. This approach ensures the adequacy of the disentanglement. The face reconstruction decoder $D_{face}$, inspired by the style transfer work SANet \cite{18huang2018multimodal, 38park2019arbitrary}. In this setup, the artifacts are used as conditions for the face reconstruction. Specifically, the artifacts are treated as the style image in style transfer tasks, while the identity information serves as the target content image. Thus, the reconstruction process is represented as follows:
\begin{equation}\label{equ11}
    \begin{gathered}
I^s_A=D_{face}(ID_A,ART_A), \\
I^s_B=D_{face}(ID_B,ART_B), \\
I^c_A=D_{face}(ID_A,ART_B), \\
I^c_B=D_{face}(ID_B,ART_A), \\
    \end{gathered}
\end{equation}
where $I^s_A$ and $I^s_B$ represent the self-reconstructed face images, while $I^c_A$ and $I^c_B$ denote the cross-reconstructed pseudo face images.

\textbf{Deepfake Detection. }With the pure identity information and pure fake features obtained from the IACC module, our classifier focuses solely on the pure artifact features. This approach eliminates the influence of identity information, thereby contributing to the development of a more generalized Deepfake detector.

\subsection{Loss Functions}
To enhance the progressive disentanglement framework for better separation of identity information and fake features, and to obtain pure identity and fake features, four distinct loss functions are designed: classification loss, reconstruction loss, Identity-Artifact Separation Contrastive loss, and Information Optimization loss to guide the IACC module. 

\textbf{Classification Loss}. Since our classifier focuses solely on pure fake features, we use the binary cross-entropy loss $L_{BCE}$ to supervise the classifier for distinguishing between real and fake faces.
\begin{equation}\label{equ12}
\begin{aligned}
    &L_{BCE}=-\frac{1}{bs\times\ (m+1)} \times\ \\ 
    &\sum_{i=1}^{b s}\left[\log \left(1-\left(p_{0}\right)_{i}\right)
    +\sum_{j=1}^{m} \log \left(\left(p_{j}\right)_{i}\right)\right], 
\end{aligned}
\end{equation}
where $bs$ represents the batch size of the input datasets, and $p_i$ is the probability assigned by the classifier that the i-th fake feature is classified as fake, where $i = \{0, 1,.., m\}$.

\textbf{Reconstruction Loss}. Our face reconstruction involves two approaches: self-reconstruction and cross-reconstruction. Self-reconstruction uses the identity information and artifact from the same person to reconstruct the corresponding face via the face reconstruction decoder $D_{face}$, while cross-reconstruction uses identity information and artifact from different people. Thus, the reconstruction loss $L^{all}_{Rec}$ is divided into self-reconstruction loss $L^{s}_{Rec}$ and cross-reconstruction loss $L^{c}_{Rec}$. The reconstruction loss ensures pixel-level consistency between the reconstructed images and the original images, while also enhancing the orthogonality between the disentangled features. The specific formula is as follows:
% \begin{equation}\label{equ13}
%     \begin{gathered}
% L_{Rec}^{s}=\left\|I_{A}-I_{A}^{s}\right\|_{1}+\left\|I_{B}-I_{B}^{s}\right\|_{1}, \\
% L_{Rec}^{c}= \left\|I_{A}-I_{A}^{c}\right\|_{1}+\left\|I_{B}-I_{B}^{c}\right\|_{1}, \\
% L_{Rec }^{all}=L_{Rec}^{s}+L_{Rec}^{c},
%     \end{gathered}
% \end{equation}
\begin{gather}
\begin{aligned}
L_{Rec}^{s}=\left\|I_{A}-I_{A}^{s}\right\|_{1}+\left\|I_{B}-I_{B}^{s}\right\|_{1}, \nonumber
\end{aligned}\\
\begin{aligned}
L_{Rec}^{c}= \left\|I_{A}-I_{A}^{c}\right\|_{1}+\left\|I_{B}-I_{B}^{c}\right\|_{1}, \nonumber 
\end{aligned}\\
\begin{aligned}\label{equ13}
L_{Rec }^{all}=L_{Rec}^{s}+L_{Rec}^{c},
\end{aligned}
\end{gather}

\textbf{Identity-Artifact Separation Contrastive Loss.} To further enhance the independence of pure artifacts, the Identity-Artifact Separation Contrastive loss is proposed, which pushes the artifacts away from the two identities present in the fake face. Considering the consistency of identity in real faces, we also need to bring the identity closer to the artifacts. Thus, the loss expression is as follows:
% \begin{equation}\label{equ14}
% \begin{gathered}
%     L_{Con}^{r}=\sum_{n=1}^{2}\left[1-\cos \left(ART_{r}, id_{r}^{n}\right)\right],\\
%     L_{Con}^{f}=\sum_{n=1}^{2}\left[\cos \left(ART_{f}, id_{f}^{n}\right)\right],\\
%     L_{Con}^{all}=L_{Con}^{r}+L_{Con}^{f},
% \end{gathered}
% \end{equation}
\begin{gather}
\begin{aligned}
L_{Con}^{r}=\sum_{n=1}^{2}\left[1-\cos \left(ART_{r}, id_{r}^{n}\right)\right], \nonumber
\end{aligned}\\
\begin{aligned}
L_{Con}^{f}=\sum_{n=1}^{2}\left[\cos \left(ART_{f}, id_{f}^{n}\right)\right], \nonumber 
\end{aligned}\\
\begin{aligned}\label{equ14}
L_{Con}^{all}=L_{Con}^{r}+L_{Con}^{f},
\end{aligned}
\end{gather}
where $L_{Con}^{r}$ and $L_{Con}^{f}$ denote the separation contrastive losses for real and fake faces, respectively. The cosine similarity $cos(x,y)$ is used to quantify the similarity between identity and artifact. Combining these two contrastive losses ensures orthogonality between identity and artifact, helping to reduce the error rate of the classifier.

\textbf{Overall Loss.} The final loss function of the training process is the weighted sum of the above three loss functions as well as the Information Optimization loss.
\begin{equation}\label{equ15}
L=\lambda_{1} L_{B C E}+\lambda_{2} L_{Rec}^{all}+\lambda_{3} L_{Con}^{all}+\lambda_{4} L_{Info},
\end{equation}
where $\lambda_1,\lambda_2,\lambda_3$ and $\lambda_4$ are hyper-parameters used to balance the overall loss.

\section{Experiments}
\subsection{Experimental Settings}
\textbf{Datasets. }To evaluate the generalization ability of the proposed method, our experiments are conducted on several standard Deepfake datasets. These datasets include FaceForensics++ (FF++) \cite{8rossler2019faceforensics++}, two versions of Celeb-DF (CDF) \cite{39li2020celeb}, DeepfakeDetection (DFD) \cite{40google2019deepfakedetection}, Deepfake Detection Challenge (DFDC) \cite{41dolhansky2020deepfake}, and a preview version of DFDC (DFDCP) \cite{42dolhansky2019deepfake}.

The FaceForensics++ (FF++) [8] is a large and widely used datasets covering more than 1.8 million fake images generated from 1000 original videos. These fake images are generated by four different face processing methods that use the same original video source, including DeepFakes (DF) \cite{4deepfakes}, Face2Face (F2F) \cite{43thies2016face2face}, FaceSwap (FS) \cite{44FaceSwap}, and NeuralTexture (NT) \cite{45thies2019deferred}.FF++ provides three different compression levels: raw, lightly compressed (c23) and heavily compressed (c40). We used the c23 version of FF++ during training, which is consistent with previous studies \cite{17yan2023ucf}. The CDF datasets consists of two versions: the Celeb-DF-v1 (CDF-v1) and the Celeb-DF-v2 (CDF-v2). The CDF-v1 contains 408 raw videos and 795 faked videos, while the CDF-v2 contains 590 real videos and 5639 Deepfake videos. DFDCP is a preview version of DFDC and contains a total of 5214 videos. DFDC is a large face-swapping datasets covering more than 110,000 video samples from 3426 actors.

\textbf{Implementation Details. }The proposed method uses the pre-trained EfficientNet-B4 \cite{34tishby2015deep} as an encoder to extract the blended identity information of the forged face. The model parameters are initialized by pre-training on the ImageNet. This method performs face extraction and alignment using DLIB \cite{46sagonas2016300}, and the size of the face images for both training and testing is scaled to $256 \times 256$. We use the Adam \cite{47kingma2014adam} optimizer for optimization, setting the learning rate to 0.0001 and the batch size to 16. Empirically, the $\lambda_1,\lambda_2,\lambda_3$ and $\lambda_4$ in the overall loss function are set to 5, 0.1, 0.5 and 0.5. To ensure fair comparisons, all experiments were conducted in DeepfakeBench \cite{21yan2023deepfakebench} and the experimental settings followed the default settings of the benchmarks. All codes are based on the PyTorch framework and trained using an NVIDIA RTX 3090.

\textbf{Evaluation Metrics. }In this paper, the default evaluation metric used is the \textbf{frame-level} Area Under the Curve (AUC) to compare the proposed method with previous competitive approaches. To facilitate comparison with other state-of-the-art detection methods, we also provide experimental results using the \textbf{video-level} AUC metric.

\subsection{Experimental Evaluation}

\begin{table}[t]
    \caption{Intra-dataset comparison results, with \textbf{frame-level AUC} as the evaluation metric.}
    \label{tab:table1}
    \normalsize
    \centering
    \renewcommand\arraystretch{1.35}
    \resizebox{\linewidth}{!}{
    \begin{tabular}{lccccc}
    \hline
    \multirow{2}{*}{Method} & \multicolumn{5}{c}{Frame-level AUC $\uparrow$} \\ \cline{2-6}
    ~                 & FF++  & DF    & F2F   & FS    & NT  \\ \hline
    FWA \cite{11li2018exposing}          & 0.877 & 0.921 & 0.900 & 0.884 & 0.812 \\
    Xception \cite{8rossler2019faceforensics++}      & 0.964 & 0.980 & 0.979 & 0.983 & 0.939 \\
    Face X-ray \cite{10li2020face}   & 0.959 & 0.979 & \underline{0.987} & 0.987 & 0.929 \\
    F3Net \cite{48qian2020thinking}        & 0.964 & 0.979 & 0.980 & 0.984 & 0.935 \\
    SRM \cite{27luo2021generalizing}          & 0.958 & 0.973 & 0.970 & 0.974 & 0.930 \\
    RECCE \cite{29cao2022end}        & 0.962 & 0.980 & 0.978 & 0.979 & 0.936 \\
    UCF \cite{17yan2023ucf}          & \underline{0.971} & \underline{0.988} & 0.984 & \underline{0.990} & \underline{0.944} \\
    Ours              & \textbf{0.982} & \textbf{0.990} & \textbf{0.989} &\textbf{0.992} & \textbf{0.962} \\ \hline
    \end{tabular}}
\end{table}

\textbf{Intra-Dataset Evaluation. }The proposed method is trained and tested on FF++ (c23) and further evaluated on the DF, F2F, FS, and NT datasets. Table \ref{tab:table1} shows the comparative results within the same dataset. Although many prior studies also use the same datasets for training and testing, differences in preprocessing steps and experimental settings may affect the fairness of the experimental results. Therefore, we implement our method in DeepfakeBench, adhering to the platform's preprocessing, experimental setup, and metric calculation methods. To ensure a fair comparison with other competitive methods, the experimental results are directly cited from DeepfakeBench \cite{21yan2023deepfakebench}. The models are trained and tested on the same datasets, using frame-level AUC as the evaluation metric. The best results are highlighted in bold, and the second-best results are underlined. As shown in Table \ref{tab:table1}, our method outperforms existing detection methods on FF++ (c23), as well as on DF, F2F, FS, and NT. For example, on the NT dataset, our method achieves a frame-level AUC of 0.962, which significantly exceeds the 0.944 AUC obtained by UCF \cite{17yan2023ucf}, a method also based on disentanglement techniques. This demonstrates a notable improvement in the detection accuracy of our proposed method on the same dataset.

\begin{table*}[ht]
    \caption{Cross-dataset comparison results, with \textbf{frame-level AUC} as the evaluation metric.}
    \label{tab:table2}
    \centering
    \renewcommand\arraystretch{1.35}
    \resizebox{0.7\linewidth}{!}{
    \begin{tabular}{lccccc}
    \hline
    \multirow{2}{*}{Method} & \multicolumn{5}{c}{Frame-level AUC $\uparrow$} \\ \cline{2-6}
    ~                       & CDF-v1  & CDF-v2    & DFD   & DFDC    & DFDCP  \\ \hline
    FWA \cite{11li2018exposing}                & 0.790 & 0.668 & 0.740 & 0.6132 & 0.638 \\
    Xception \cite{8rossler2019faceforensics++}            & 0.779 & 0.737 & \underline{0.816} & 0.708 & 0.737 \\
    EfficientNet-B4 \cite{33tan2019efficientnet}    & 0.791 & 0.749 & 0.815 & 0.696 & 0.728 \\
    Face X-ray \cite{10li2020face}         & 0.709 & 0.679 & 0.766 & 0.633 & 0.694 \\
    F3Net \cite{48qian2020thinking}             & 0.777 & 0.735 & 0.798 & 0.702 & 0.735 \\
    SPSL \cite{49liu2021spatial}              & \underline{0.815} & 0.765 & 0.812 & 0.704 & 0.741  \\
    SRM \cite{27luo2021generalizing}                & 0.793 & 0.755 & 0.812 & 0.700 & 0.741 \\
    CORE \cite{50ni2022core}               & 0.782 & 0.743 & 0.802 & 0.705 & 0.734 \\
    RECCE \cite{29cao2022end}              & 0.768 & 0.732 & 0.812 & 0.713 & 0.742 \\
    IID \cite{31huang2023implicit}                & 0.758 & \underline{0.769} & 0.794 & 0.695 & - \\
    UCF \cite{17yan2023ucf}                & 0.779 & 0.753 & 0.807 & \underline{0.719} & \underline{0.759} \\
    Ours                    & \textbf{0.825} & \textbf{0.813} & \textbf{0.856} &\textbf{0.760} & \textbf{0.778} \\ \hline
    \end{tabular}}
\end{table*}

\textbf{Cross-Dataset Evaluation. }To validate the superior generalization ability of our proposed method over existing detection methods on unknown datasets, cross-dataset generalization experiments are conducted. We first train the model on the FF++ (c23) dataset and then test it on CDF-v1, CDF-v2, DFD, DFDC, and DFDCP. To ensure fairness in comparison, all experimental results for comparison methods come from DeepfakeBench \cite{21yan2023deepfakebench}, and our method is also evaluated within DeepfakeBench. All detectors are trained on FF++ (c23) and evaluated on other datasets, with frame-level AUC as the evaluation metric. The best results are highlighted in bold, and the second-best results are underlined. The cross-dataset comparison results are shown in Table \ref{tab:table2}, where our method achieves the best results across all five benchmark datasets, demonstrating the superior generalization ability of the proposed progressive disentanglement framework. The UCF \cite{17yan2023ucf}, another disentanglement-based method, only achieves second-best results on DFDC and DFDCP, indicating that its disentanglement approach inspired by style transfer task is not fully suitable for Deepfake detection tasks, thus compromising the reliability of its disentanglement method. In contrast, our disentanglement framework, designed based on artifact generation mechanisms, aims to more accurately separate identity and artifacts. By incorporating the IACC module guided by Information Bottleneck (IB) theory, the potential correlation between identity and artifacts is further eliminated, and pure identity information and artifacts are obtained. Our detector focuses solely on pure artifact features, avoiding the interference of identity information. Consequently, our detector does not overfit the global features of the training dataset, enhancing the generalization ability. Our experimental results on the five benchmark datasets improve over sub-optimal results, reaching 0.825, 0.813, 0.856, 0.760, and 0.778, respectively. On CDF-v2, DFD, and DFDC, the sub-optimal results are improved by 0.044, 0.04, and 0.041, respectively.. Overall, the experimental results prove that our method has high detection accuracy and strong generalization ability. 

\begin{table}[t]
    \caption{Comparison results with the latest SOTA methods, with \textbf{video-level AUC} as the evaluation metric.}
    \label{tab:table3}
    \centering
    \renewcommand\arraystretch{1.35}
    \resizebox{\linewidth}{!}{
    \begin{tabular}{lccc}
    \hline
    \multirow{2}{*}{Method} &\multirow{2}{*}{Publication} & \multicolumn{2}{c}{Video-level AUC $\uparrow$} \\ \cline{3-4}
    ~                      & ~         & CDF-v2  & DFDC  \\ \hline
    LipForensice \cite{24haliassos2021lips}      & CVPR’21   & 0.824   & 0.735  \\
    FTCN \cite{26zheng2021exploring}              & ICCV’21   & \textbf{0.869} & 0.740 \\
    DCL \cite{28sun2022dual}               & AAAI’22   & 0.823   & \underline{0.767}  \\
    HCIL \cite{51gu2022hierarchical}              & ECCV’22   & 0.790   & 0.692  \\
    Liang et al. \cite{15liang2022exploring}      & ECCV’22   & 0.824   & -     \\
    RealForensics \cite{52haliassos2022leveraging}     & CVPR’22   & 0.857   & 0.759  \\
    ITSNet \cite{58wu2023interactive}          & TCSVT'23  & 0.860   & - \\
    SFDDG \cite{53wang2023dynamic}             & CVPR’23   & 0.758   & 0.736  \\
    FADE \cite{54tan2023deepfake}              & AAAI’23   & 0.775   & -  \\
    Zhang et al. \cite{59zhang2024face}        & TCSVT'24  & -       & 0.713 \\
    Ours                   & -         & \underline{0.864} & \textbf{0.776} \\ \hline
    \end{tabular}}
\end{table}

\textbf{Comparison with Other SOTA Methods. }Given that our method uses video frames as input, we compare it not only to the frame-level AUC results of the detector implemented in DeepfakeBench but also to other state-of-the-art (SOTA) methods using video-level AUC. The data is cited directly from their original papers of these detectors. Although comparing with SOTA methods is challenging, our method still performs well. All detectors are trained on FF++ (c23), with video-level AUC as the evaluation metric. The best results are highlighted in bold, and the second-best results are underlined. The comparison results are shown in Table \ref{tab:table3}. Our method achieves second-best results on the CDF-v2 dataset and the best performance on the DFDC dataset. Liang et al. \cite{15liang2022exploring} also employed Deepfake detection methods based on disentanglement techniques. In comparison, our method achieved a video-level AUC of 0.864 on the CDF-v2 dataset, which is significantly better than their results, demonstrating a marked improvement in Deepfake detection performance. Compared to the RealForensics \cite{52haliassos2022leveraging} method, the detection result of our method on the DFDC dataset by 0.017, further proving the effective generalization ability of our approach.

Through comparing with other competitive detection methods, we demonstrate that the proposed method not only enhances the detection accuracy but also improves the generalization ability when facing unknown datasets. Compared to disentanglement-based methods (such as Liang et al. \cite{15liang2022exploring} and UCF \cite{17yan2023ucf}), our method addresses the generalization issue more effectively. This improvement is the result of  the disentanglement strategy being designed based on the inherent characteristics of Deepfake artifacts, ensuring its reliability. Additionally, the well-designed IACC module effectively mitigates the impact of identity information on fake detection.

\subsection{Ablation Study}
To evaluate the impact of the proposed progressive disentanglement framework, the IACC module based on Information Bottleneck (IB) theory, and the Identity-Artifact Separation Contrastive loss $L_{Con}$ on the detector, the ablation experiments are conducted to study their effects on model generalization. The EfficientNet-B4 (EFN) \cite{33tan2019efficientnet} model is used as the baseline and evaluate the following variants: an improved EFN model with the progressive disentanglement (PD) framework, an improved EFN model with the IACC module guided by the Information Optimization loss $L_{Info}$, and the complete model with the additional Identity-Artifact Separation Contrastive loss $L_{Con}$. Specifically,  all variants are trained on the FF++ (c23) dataset and tested on FF++, CDF-v1, DFD, and DFDC.

The ablation study results are shown in Table \ref{tab:table4}. It can be observed that all variants achieve higher detection accuracy compared to the baseline EfficientNet-B4 model, indicating that the proposed method effectively enhances model performance. Furthermore, the performance improves progressively as more components are added to the model. Firstly, the progressive disentanglement framework is introduced, which significantly enhances the generalization ability of the improved EFN model on the test datasets. This demonstrates that our framework effectively combines coarse-grained and fine-grained strategies to progressively separate identity information and artifacts, ensuring the reliability of the disentanglement approach. Next, incorporating the IACC module based on Information Bottleneck (IB) theory further improves the detection performance by obtaining pure identity and artifact features, highlighting the importance of thoroughly exploring and reducing the potential correlation between identity and artifacts. Finally, the addition of the Identity-Artifact Separation Contrastive loss $L_{Con}$ further refines the separation of pure identity information and artifacts, as real faces have a consistent identity, necessitating proper alignment between the identity information and pseudo artifacts. Overall, our complete model achieves the best detection performance across the entire dataset. Through the ablation study, the effectiveness of each component proposed for the artifact generation mechanism is validated.

\begin{table}[htp]
    \caption{Ablation study for all variants, with \textbf{frame-level AUC} as the evaluation metric.}
    \normalsize
    \label{tab:table4}
    \renewcommand\arraystretch{1.35}
    \resizebox{\linewidth}{!}{
    \centering
    \begin{tabular}{lccccc}
    \hline
     \multirow{2}{*}{Method} & \multicolumn{5}{c}{Frame-level AUC $\uparrow$} \\
     \cline{2-6}
     ~                      & FF++             & CDF-v2  & DFD   & DFDC  & Avg. \\ \hline
    EFN                     & 0.957            & 0.791   & 0.815 & 0.696 & 0.815\\
    EFN+PD                  & 0.975            & 0.794   & 0.826 & 0.745 & 0.835\\
    EFN+PD+IACC             & 0.981            & 0.815   & 0.840 &\textbf{0.761} & 0.849\\
    EFN+PD+IACC+$L_{Con}$   & \textbf{0.983}   &\textbf{0.825}  &\textbf{0.856}  & 0.760  & \textbf{0.856}\\
    \hline
    \end{tabular}}
\end{table}

\subsection{Visualization}
\begin{figure*}[ht]
    \centering
    \subfigure[Non-processed]{
        \includegraphics[width=0.3\linewidth]{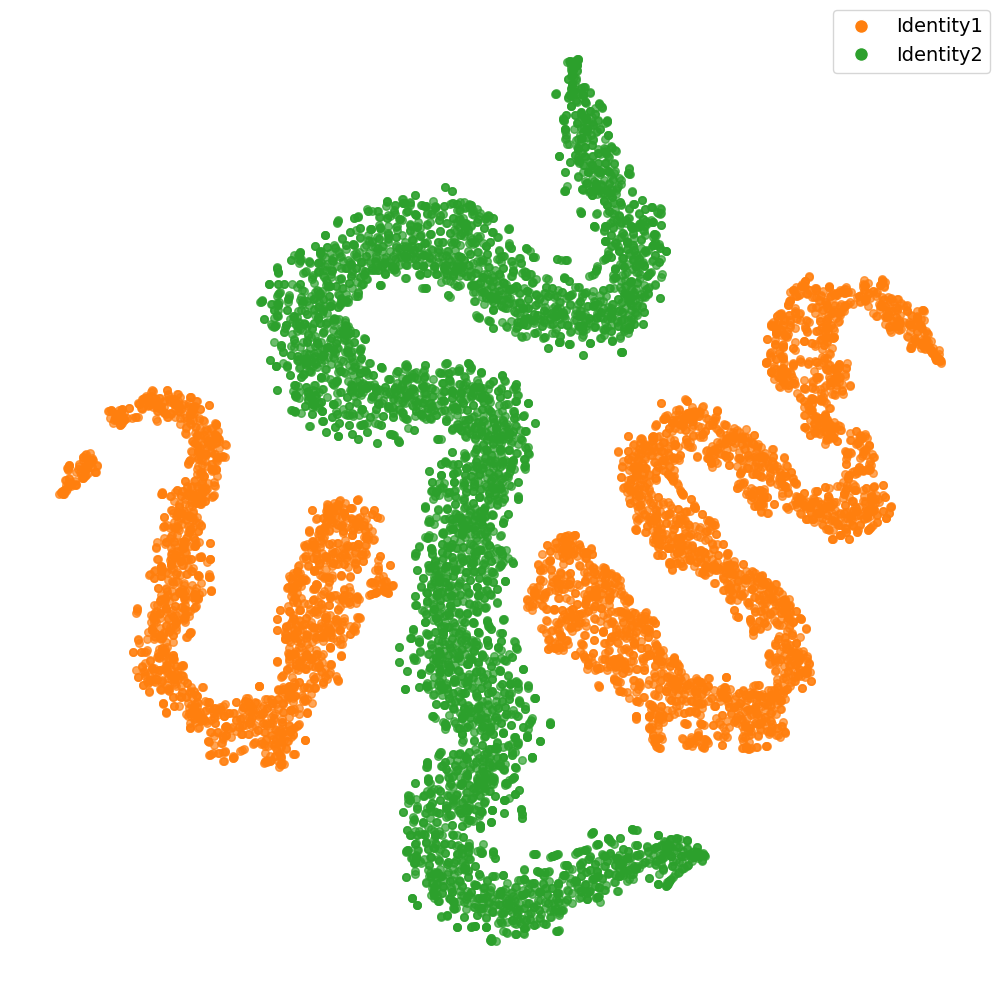}
        \label{fig:subfig3_a}
    }
    \hfill
    \subfigure[PD]{
        \includegraphics[width=0.3\linewidth]{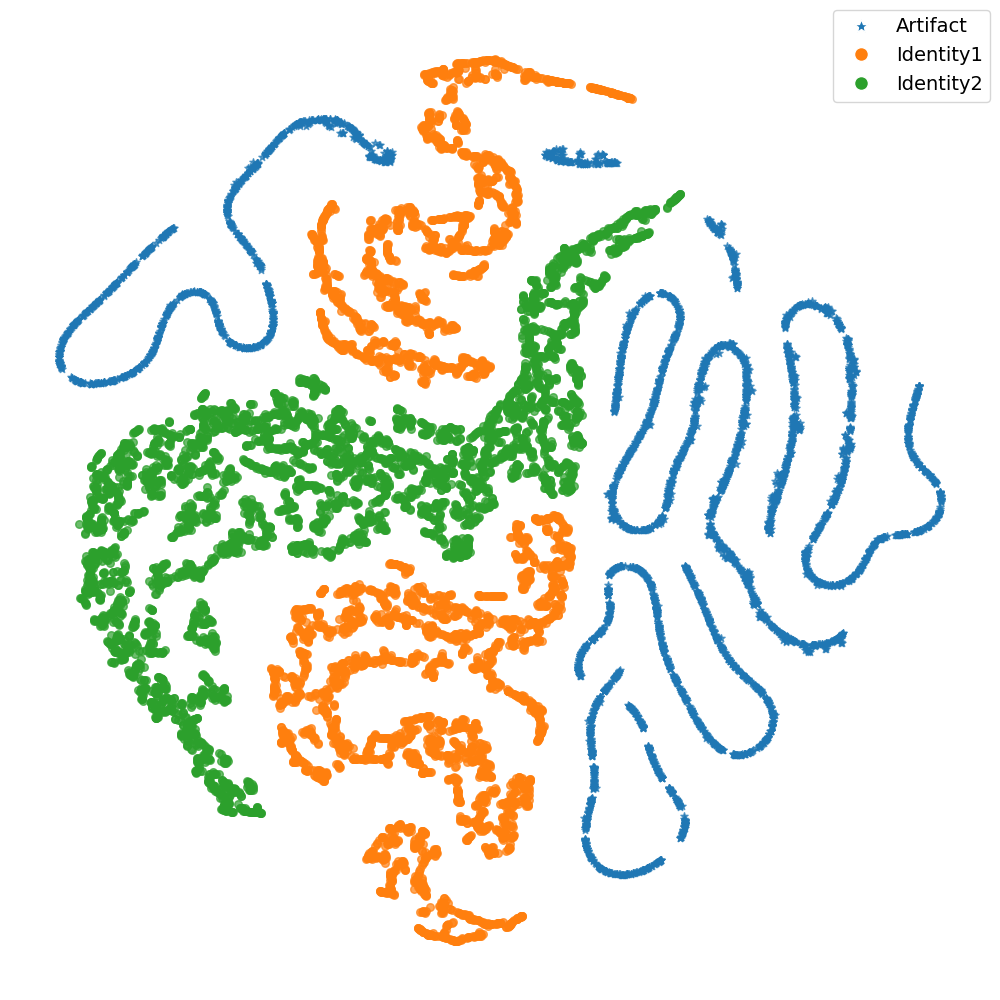}
        \label{fig:subfig3_b}
    }
    \hfill
    \subfigure[Ours]{
        \includegraphics[width=0.3\linewidth]{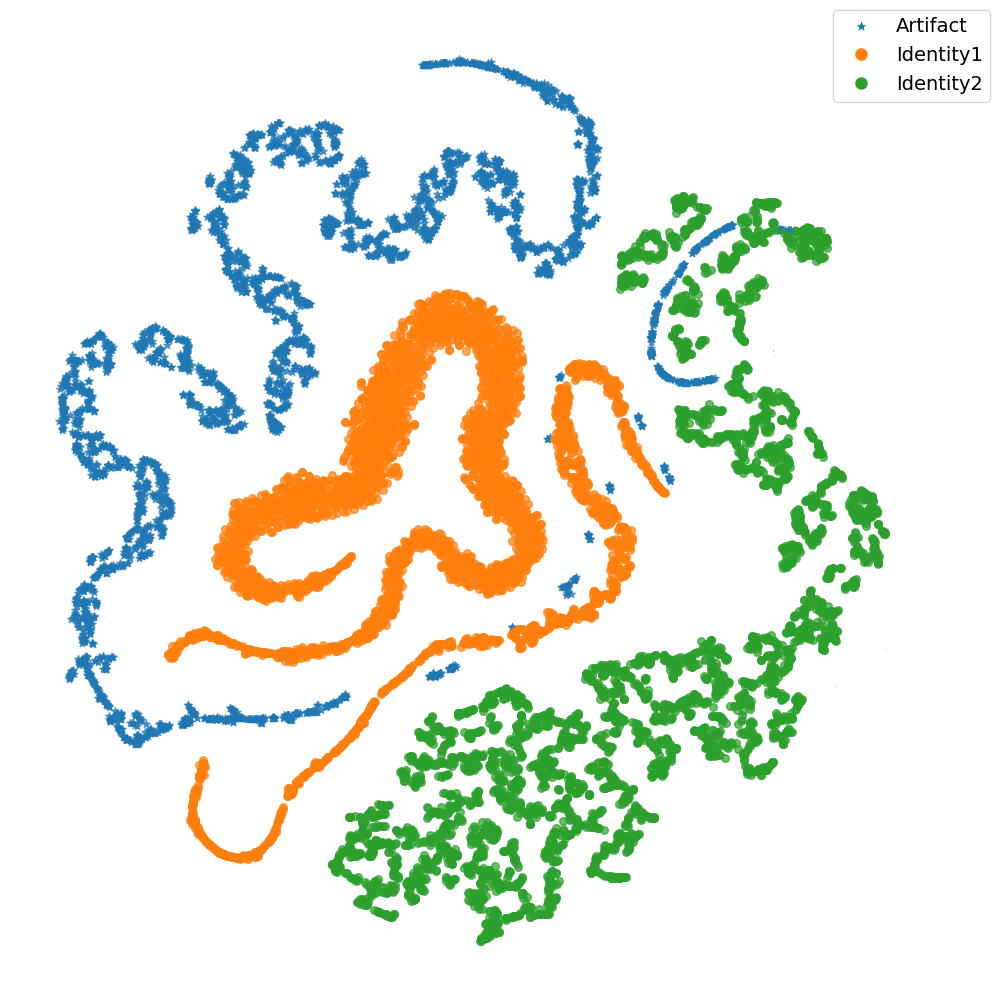}
        \label{fig:subfig3_c}
    }
    \caption{The t-SNE visualization illustrates the feature distributions of identity information and artifacts. 
    (a) t-SNE visualization of unprocessed mixed identity information. 
    (b) t-SNE visualization of identity information and artifacts after applying the basic progressive disentanglement framework (w/o IACC+$L_{Con}$). 
    (c) t-SNE visualization of the pure identity information and pure artifacts obtained using our proposed method. }
    \label{fig:fig_3}
\end{figure*}

\textbf{Visualization of Identity and Artifact Distribution. }To more clearly demonstrate the disentanglement capability of our proposed method, the t-SNE \cite{55van2008visualizing} is used to visualize the feature distributions of identity information and artifacts. This experiment is trained and tested on the FF++ (c23) dataset. First, 5000 samples are randomly selected from the FF++ (c23) dataset to visualize the distribution of the unprocessed mixed identity information in the feature space. As shown in Fig. \ref{fig:fig_3}\subref{fig:subfig3_a}, the distribution areas of the two types of identity information in the feature space are not entirely independent, indicating that there is a certain degree of coupling between them due to the presence of artifacts.

To address this issue, a progressive disentanglement method based on the artifact generation mechanism is proposed, aiming to separate artifacts from the mixed identity information. Fig. \ref{fig:fig_3}\subref{fig:subfig3_b} shows the feature distribution of identity information and artifacts under our basic progressive disentanglement framework (excluding the IACC module and all optimization loss functions). Although the separation effect is improved, there is still some overlap between the distributions of artifacts and identity information. To further reduce the potential correlation between identity information and artifacts, the IACC module based on Information Bottleneck theory and the Identity-Artifact Separation Contrastive loss are introduced into the framework to extract pure identity information and artifacts. Similarly, another 5000 samples are selected from the FF++ (c23) dataset to visualize the distribution of pure identity information and artifacts in the feature space. The results, as shown in Fig. \ref{fig:fig_3}\subref{fig:subfig3_c}, demonstrate that the distributions of artifacts and identity information in the feature space are significantly more independent, proving that our method effectively separates artifacts from identity information, thereby significantly improving the accuracy of model detection. 

% figure 3
% \begin{figure*}[ht]
%     \centering
%     % Subfigure (a)
%     \subfloat[Non-processed]{\includegraphics[width=0.32\linewidth]{fig3_a_.png}
%     \label{fig:subfig3_a}}
%     \hfil
%     % Subfigure (b)
%     \subfloat[PD]{\includegraphics[width=0.32\linewidth]{fig3_b_.png}
%     \label{fig:subfig3_b}}
%     \hfil
%     % Subfigure (c)
%     \subfloat[Ours]{\includegraphics[width=0.32\linewidth]{fig3_c_.png}
%     \label{fig:subfig3_c}}
%     \hfil
%     \caption{The t-SNE visualization illustrates the feature distributions of identity information and artifacts. 
%     (a) t-SNE visualization of unprocessed mixed identity information. 
%     (b) t-SNE visualization of identity information and artifacts after applying the basic progressive disentanglement framework (w/o IACC+$L_{Con}$). 
%     (c) t-SNE visualization of the pure identity information and pure artifacts obtained using our proposed method. }
%     \label{fig:fig_3}
% \end{figure*}

\begin{figure}[t]
    \centering
    \subfigure[EfficientNet-B4]{
        \includegraphics[width=0.4\linewidth]{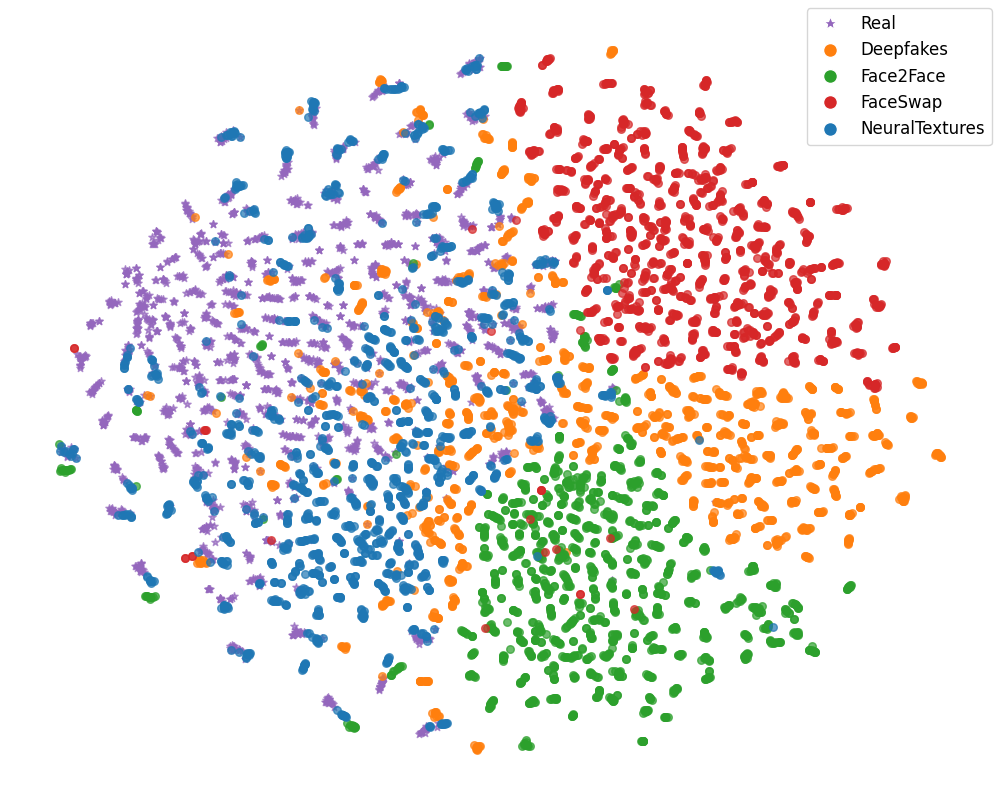}
        \label{fig:subfig4_a}
    }
    \hfill
    \subfigure[Ours]{
        \includegraphics[width=0.4\linewidth]{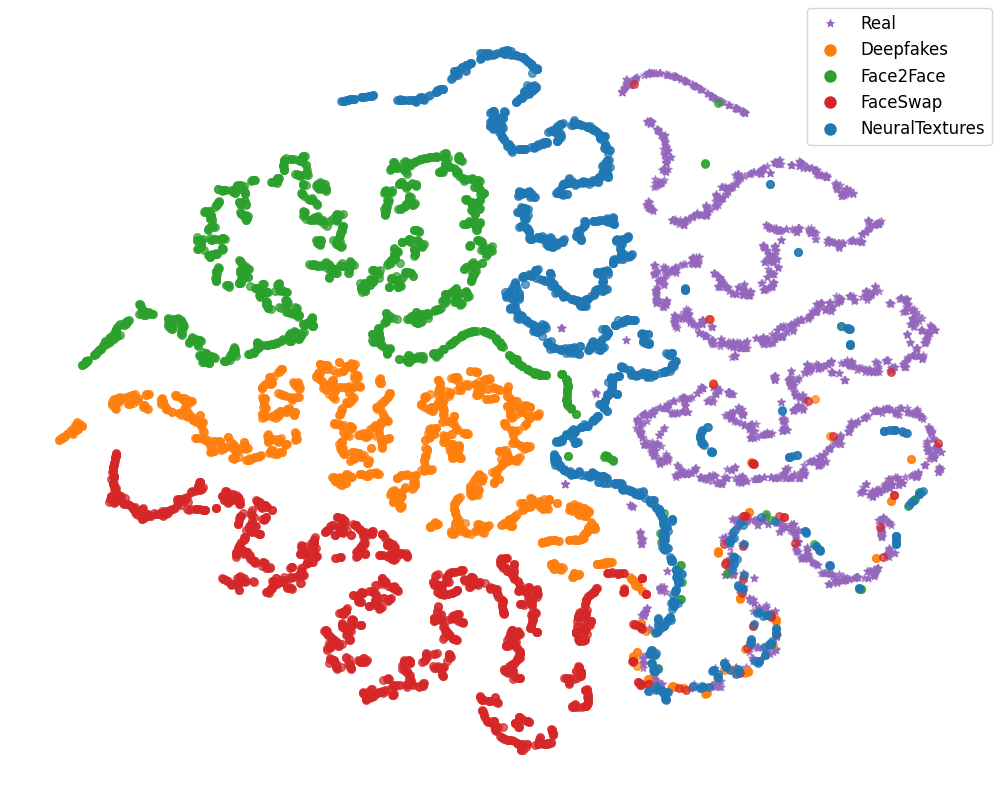}
        \label{fig:subfig4_b}
    }
    \caption{The t-SNE visualization of the multi-domain feature distribution extracted from the FF++ (c23) dataset by EfficientNet-B4 and our proposed method.}
    \label{fig:fig_4}
\end{figure}

\textbf{Visualization of multi-domain datasets distribution. }To more intuitively demonstrate the generalization ability of our method, we used t-SNE to visualize the feature distribution on the FF++ (c23) \cite{8rossler2019faceforensics++} dataset. This dataset includes fake faces generated by four different forgery methods as well as original real faces. We trained and tested both the EfficientNet-B4 model and our proposed progressive disentanglement method on the FF++ (c23) dataset. Fig. \ref{fig:fig_4}\subref{fig:subfig4_a} shows the feature distribution learned by the EfficientNet-B4 model, which aligns with the results of the quantitative experiments. It can be seen that the feature distribution areas of DF, F2F, and FS are more independent compared to NT, with clearer decision boundaries. However, the distribution of NT is blended with real faces. This is because the NT method learns neural textures for face reconstruction, making it more difficult to detect the forgery, which poses a significant challenge.

% figure 4
% \begin{figure}[t]
%     \centering
%     % Subfigure (a)
%     \subfloat[EfficientNet-B4]{\includegraphics[width=0.5\linewidth]{fig4_a_.png}
%     \label{fig:subfig4_a}}
%     % \hfil
%     % Subfigure (b)
%     \subfloat[Ours]{\includegraphics[width=0.5\linewidth]{fig4_b_.png}
%     \label{fig:subfig4_b}}
%     % \hfil
%     \caption{The t-SNE visualization of the multi-domain feature distribution extracted from the FF++ (c23) dataset by EfficientNet-B4 and our proposed method.}
%     \label{fig:fig_4}
% \end{figure}

As expected, Fig. \ref{fig:fig_4}\subref{fig:subfig4_b} shows that the features extracted by our method are significantly more separated, with more robust decision boundaries. Furthermore, the feature representation of each class is more compact, further enhancing the robustness of the method. Notably, our method achieved superior detection results on the NT dataset, indicating that it can learn more generalizable feature representations. This is due to our method, which is based on the artifact generation mechanism. By using the IACC module, it effectively compresses the potential correlation between identity information and artifacts, resulting in purer artifact feature representations, rather than overfitting to specific forgery methods. As a result, our method demonstrates stronger detection performance and generalization ability.

\begin{figure}[h]
    \centering
    \includegraphics[width=1\linewidth]{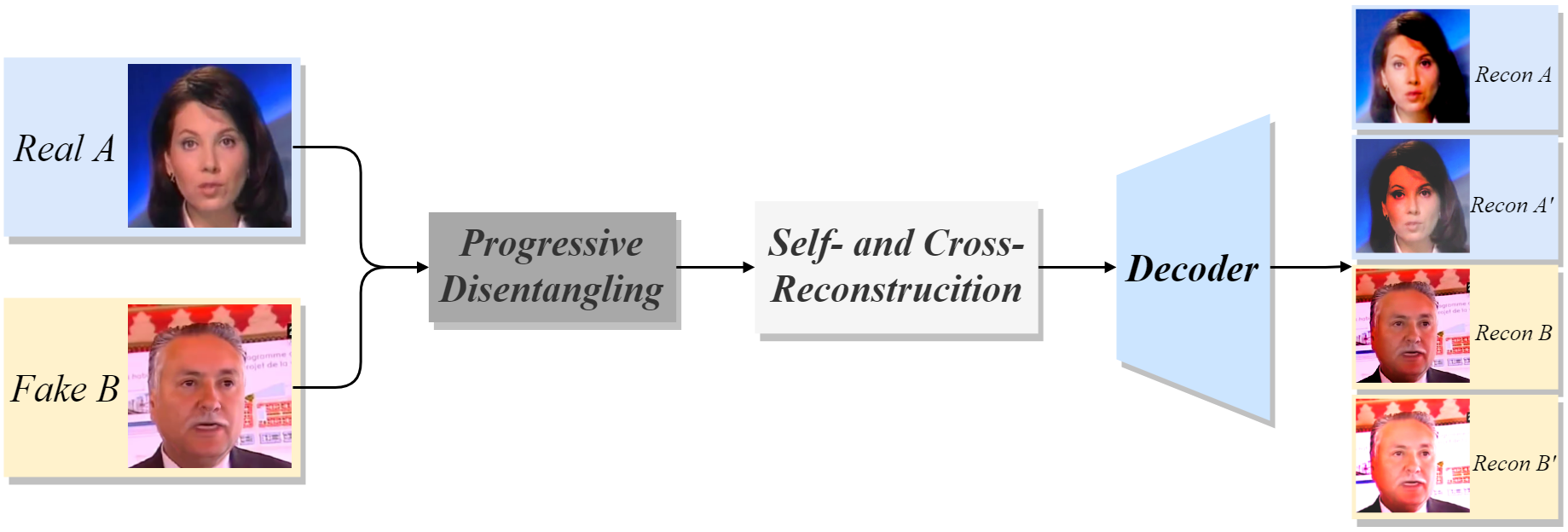}
    \caption{Visualization of face self-reconstruction and cross-reconstruction in the progressive disentanglement framework.}
    \label{fig:fig_5}
\end{figure}

\textbf{Visualization of Reconstructed Faces. }In the proposed progressive disentanglement method, both identity self-reconstruction and cross-reconstruction are performed to ensure that the identity information and artifacts are orthogonal after disentanglement. The visualization results are shown in Fig. \ref{fig:fig_5}. The essence of the disentanglement framework is to separate identity information from artifacts, so the visual appearance of the reconstructed image is more likely to emphasize identity information. However, the artifacts are visually represented in the images. For example, the reconstructed fake face $A'$ in Fig. \ref{fig:fig_5} shows red areas not present in the original image. This could be due to the reddish skin tone of the fake face $B$, which causes the artifacts from $B$ to introduce a red hue to the facial area in $A'$. Therefore, the effective separation of identity information and artifacts is realized by the disentanglement framework. 

% figure5

\section{Conclusion}
In this paper, a novel Deepfake detection method based on progressive disentanglement and purification of blended identities is proposd to capture and separate artifact features in fake faces more accurately. The disentanglement framework leverages the artifact generation mechanism to ensure the reliability of the process. Initially, the face is coarsely disentangled into a pair of blended identities to address the identity inconsistencies commonly found in Deepfakes. These identities are then finely separated into non-pure identity information and blended artifacts. To further explore and mitigate the potential correlations between identity and artifacts, the Identity-Artifact Correlation Compression (IACC) module is introduced, which is grounded in Information Bottleneck (IB) theory, to extract pure identity information and artifacts. In addition, an Identity-Artifact Separation Contrastive loss function is designed to enhance the independence of artifact features. Ultimately, the classifier focuses exclusively on pure artifact features, eliminating the influence of identity information on detection results. Extensive experiments across six benchmark datasets demonstrate that the proposed method significantly outperforms existing state-of-the-art techniques, exhibiting stronger detection performance and generalization capabilities. We hope this research encourages further exploration of disentanglement techniques in the field of Deepfake detection.

\section{Acknowledge}
This work was supported in part by the National Key Research and Development Program of China under Grant (2023YFF1105102, 2023YFF1105105, SQ2023YFF1100111), the National Natural Science Foundation of China under Grant 61772237, the Joint Fund of Ministry of Education for Equipment Pre-research under Grant 8091B042236 and Jiangnan University graduate research and practice innovation project under Grant 2050205.


\bibliographystyle{IEEEtran}
\bibliography{References.bib}

\begin{thebibliography}{10}
\providecommand{\url}[1]{#1}
\csname url@samestyle\endcsname
\providecommand{\newblock}{\relax}
\providecommand{\bibinfo}[2]{#2}
\providecommand{\BIBentrySTDinterwordspacing}{\spaceskip=0pt\relax}
\providecommand{\BIBentryALTinterwordstretchfactor}{4}
\providecommand{\BIBentryALTinterwordspacing}{\spaceskip=\fontdimen2\font plus
\BIBentryALTinterwordstretchfactor\fontdimen3\font minus \fontdimen4\font\relax}
\providecommand{\BIBforeignlanguage}[2]{{%
\expandafter\ifx\csname l@#1\endcsname\relax
\typeout{** WARNING: IEEEtran.bst: No hyphenation pattern has been}%
\typeout{** loaded for the language `#1'. Using the pattern for}%
\typeout{** the default language instead.}%
\else
\language=\csname l@#1\endcsname
\fi
#2}}
\providecommand{\BIBdecl}{\relax}
\BIBdecl

\bibitem{1bitouk2008face}
D.~Bitouk, N.~Kumar, S.~Dhillon, P.~Belhumeur, and S.~K. Nayar, ``Face swapping: automatically replacing faces in photographs,'' in \emph{ACM SIGGRAPH 2008 papers}, 2008, pp. 1--8.

\bibitem{3korshunova2017fast}
I.~Korshunova, W.~Shi, J.~Dambre, and L.~Theis, ``Fast face-swap using convolutional neural networks,'' in \emph{Proceedings of the IEEE international conference on computer vision}, 2017, pp. 3677--3685.

\bibitem{4deepfakes}
\BIBentryALTinterwordspacing
Deepfakes github. Accessed: Oct. 2023. [Online]. Available: \url{https://github.com/deepfakes/faceswap}
\BIBentrySTDinterwordspacing

\bibitem{6lyu2020deepfake}
S.~Lyu, ``Deepfake detection: Current challenges and next steps,'' in \emph{2020 IEEE international conference on multimedia \& expo workshops (ICMEW)}.\hskip 1em plus 0.5em minus 0.4em\relax IEEE, 2020, pp. 1--6.

\bibitem{7afchar2018mesonet}
D.~Afchar, V.~Nozick, J.~Yamagishi, and I.~Echizen, ``Mesonet: a compact facial video forgery detection network,'' in \emph{2018 IEEE international workshop on information forensics and security (WIFS)}.\hskip 1em plus 0.5em minus 0.4em\relax IEEE, 2018, pp. 1--7.

\bibitem{8rossler2019faceforensics++}
A.~Rossler, D.~Cozzolino, L.~Verdoliva, C.~Riess, J.~Thies, and M.~Nie{\ss}ner, ``Faceforensics++: Learning to detect manipulated facial images,'' in \emph{Proceedings of the IEEE/CVF international conference on computer vision}, 2019, pp. 1--11.

\bibitem{10li2020face}
L.~Li, J.~Bao, T.~Zhang, H.~Yang, D.~Chen, F.~Wen, and B.~Guo, ``Face x-ray for more general face forgery detection,'' in \emph{Proceedings of the IEEE/CVF conference on computer vision and pattern recognition}, 2020, pp. 5001--5010.

\bibitem{11li2018exposing}
Y.~Li and S.~Lyu, ``Exposing deepfake videos by detecting face warping artifacts,'' \emph{arXiv preprint arXiv:1811.00656}, 2018.

\bibitem{13zhao2021multi}
H.~Zhao, W.~Zhou, D.~Chen, T.~Wei, W.~Zhang, and N.~Yu, ``Multi-attentional deepfake detection,'' in \emph{Proceedings of the IEEE/CVF conference on computer vision and pattern recognition}, 2021, pp. 2185--2194.

\bibitem{14dong2023implicit}
S.~Dong, J.~Wang, R.~Ji, J.~Liang, H.~Fan, and Z.~Ge, ``Implicit identity leakage: The stumbling block to improving deepfake detection generalization,'' in \emph{Proceedings of the IEEE/CVF Conference on Computer Vision and Pattern Recognition}, 2023, pp. 3994--4004.

\bibitem{15liang2022exploring}
J.~Liang, H.~Shi, and W.~Deng, ``Exploring disentangled content information for face forgery detection,'' in \emph{European Conference on Computer Vision}.\hskip 1em plus 0.5em minus 0.4em\relax Springer, 2022, pp. 128--145.

\bibitem{16hu2021improving}
J.~Hu, S.~Wang, and X.~Li, ``Improving the generalization ability of deepfake detection via disentangled representation learning,'' in \emph{2021 IEEE International Conference on Image Processing (ICIP)}.\hskip 1em plus 0.5em minus 0.4em\relax IEEE, 2021, pp. 3577--3581.

\bibitem{56li2022artifacts}
X.~Li, R.~Ni, P.~Yang, Z.~Fu, and Y.~Zhao, ``Artifacts-disentangled adversarial learning for deepfake detection,'' \emph{IEEE Transactions on Circuits and Systems for Video Technology}, vol.~33, no.~4, pp. 1658--1670, 2022.

\bibitem{17yan2023ucf}
Z.~Yan, Y.~Zhang, Y.~Fan, and B.~Wu, ``Ucf: Uncovering common features for generalizable deepfake detection,'' in \emph{Proceedings of the IEEE/CVF International Conference on Computer Vision}, 2023, pp. 22\,412--22\,423.

\bibitem{18huang2018multimodal}
X.~Huang, M.-Y. Liu, S.~Belongie, and J.~Kautz, ``Multimodal unsupervised image-to-image translation,'' in \emph{Proceedings of the European conference on computer vision (ECCV)}, 2018, pp. 172--189.

\bibitem{21yan2023deepfakebench}
Z.~Yan, Y.~Zhang, X.~Yuan, S.~Lyu, and B.~Wu, ``Deepfakebench: a comprehensive benchmark of deepfake detection,'' in \emph{Proceedings of the 37th International Conference on Neural Information Processing Systems}, 2023, pp. 4534--4565.

\bibitem{22li2018ictu}
Y.~Li, M.-C. Chang, and S.~Lyu, ``In ictu oculi: Exposing ai created fake videos by detecting eye blinking,'' in \emph{2018 IEEE International workshop on information forensics and security (WIFS)}.\hskip 1em plus 0.5em minus 0.4em\relax Ieee, 2018, pp. 1--7.

\bibitem{23yang2019exposing}
X.~Yang, Y.~Li, and S.~Lyu, ``Exposing deep fakes using inconsistent head poses,'' in \emph{ICASSP 2019-2019 IEEE International Conference on Acoustics, Speech and Signal Processing (ICASSP)}.\hskip 1em plus 0.5em minus 0.4em\relax IEEE, 2019, pp. 8261--8265.

\bibitem{24haliassos2021lips}
A.~Haliassos, K.~Vougioukas, S.~Petridis, and M.~Pantic, ``Lips don't lie: A generalisable and robust approach to face forgery detection,'' in \emph{Proceedings of the IEEE/CVF conference on computer vision and pattern recognition}, 2021, pp. 5039--5049.

\bibitem{57wang2023exploiting}
H.~Wang, Z.~Liu, and S.~Wang, ``Exploiting complementary dynamic incoherence for deepfake video detection,'' \emph{IEEE Transactions on Circuits and Systems for Video Technology}, vol.~33, no.~8, pp. 4027--4040, 2023.

\bibitem{25shiohara2022detecting}
K.~Shiohara and T.~Yamasaki, ``Detecting deepfakes with self-blended images,'' in \emph{Proceedings of the IEEE/CVF Conference on Computer Vision and Pattern Recognition}, 2022, pp. 18\,720--18\,729.

\bibitem{26zheng2021exploring}
Y.~Zheng, J.~Bao, D.~Chen, M.~Zeng, and F.~Wen, ``Exploring temporal coherence for more general video face forgery detection,'' in \emph{Proceedings of the IEEE/CVF international conference on computer vision}, 2021, pp. 15\,044--15\,054.

\bibitem{27luo2021generalizing}
Y.~Luo, Y.~Zhang, J.~Yan, and W.~Liu, ``Generalizing face forgery detection with high-frequency features,'' in \emph{Proceedings of the IEEE/CVF conference on computer vision and pattern recognition}, 2021, pp. 16\,317--16\,326.

\bibitem{60hu2021detecting}
J.~Hu, X.~Liao, W.~Wang, and Z.~Qin, ``Detecting compressed deepfake videos in social networks using frame-temporality two-stream convolutional network,'' \emph{IEEE Transactions on Circuits and Systems for Video Technology}, vol.~32, no.~3, pp. 1089--1102, 2021.

\bibitem{28sun2022dual}
K.~Sun, T.~Yao, S.~Chen, S.~Ding, J.~Li, and R.~Ji, ``Dual contrastive learning for general face forgery detection,'' in \emph{Proceedings of the AAAI Conference on Artificial Intelligence}, vol.~36, no.~2, 2022, pp. 2316--2324.

\bibitem{29cao2022end}
J.~Cao, C.~Ma, T.~Yao, S.~Chen, S.~Ding, and X.~Yang, ``End-to-end reconstruction-classification learning for face forgery detection,'' in \emph{Proceedings of the IEEE/CVF Conference on Computer Vision and Pattern Recognition}, 2022, pp. 4113--4122.

\bibitem{58wu2023interactive}
J.~Wu, B.~Zhang, Z.~Li, G.~Pang, Z.~Teng, and J.~Fan, ``Interactive two-stream network across modalities for deepfake detection,'' \emph{IEEE Transactions on Circuits and Systems for Video Technology}, vol.~33, no.~11, pp. 6418--6430, 2023.

\bibitem{31huang2023implicit}
B.~Huang, Z.~Wang, J.~Yang, J.~Ai, Q.~Zou, Q.~Wang, and D.~Ye, ``Implicit identity driven deepfake face swapping detection,'' in \emph{Proceedings of the IEEE/CVF conference on computer vision and pattern recognition}, 2023, pp. 4490--4499.

\bibitem{59zhang2024face}
D.~Zhang, J.~Chen, X.~Liao, F.~Li, J.~Chen, and G.~Yang, ``Face forgery detection via multi-feature fusion and local enhancement,'' \emph{IEEE Transactions on Circuits and Systems for Video Technology}, 2024.

\bibitem{32bengio2013representation}
Y.~Bengio, A.~Courville, and P.~Vincent, ``Representation learning: A review and new perspectives,'' \emph{IEEE transactions on pattern analysis and machine intelligence}, vol.~35, no.~8, pp. 1798--1828, 2013.

\bibitem{33tan2019efficientnet}
M.~Tan and Q.~Le, ``Efficientnet: Rethinking model scaling for convolutional neural networks,'' in \emph{International conference on machine learning}.\hskip 1em plus 0.5em minus 0.4em\relax PMLR, 2019, pp. 6105--6114.

\bibitem{20tishby2000information}
N.~Tishby, F.~C. Pereira, and W.~Bialek, ``The information bottleneck method,'' \emph{arXiv preprint physics/0004057}, 2000.

\bibitem{34tishby2015deep}
N.~Tishby and N.~Zaslavsky, ``Deep learning and the information bottleneck principle,'' in \emph{2015 ieee information theory workshop (itw)}.\hskip 1em plus 0.5em minus 0.4em\relax IEEE, 2015, pp. 1--5.

\bibitem{35huang2019ccnet}
Z.~Huang, X.~Wang, L.~Huang, C.~Huang, Y.~Wei, and W.~Liu, ``Ccnet: Criss-cross attention for semantic segmentation,'' in \emph{Proceedings of the IEEE/CVF international conference on computer vision}, 2019, pp. 603--612.

\bibitem{36smilkov2017smoothgrad}
D.~Smilkov, N.~Thorat, B.~Kim, F.~Vi{\'e}gas, and M.~Wattenberg, ``Smoothgrad: removing noise by adding noise,'' \emph{arXiv preprint arXiv:1706.03825}, 2017.

\bibitem{37karras2020analyzing}
T.~Karras, S.~Laine, M.~Aittala, J.~Hellsten, J.~Lehtinen, and T.~Aila, ``Analyzing and improving the image quality of stylegan,'' in \emph{Proceedings of the IEEE/CVF conference on computer vision and pattern recognition}, 2020, pp. 8110--8119.

\bibitem{38park2019arbitrary}
D.~Y. Park and K.~H. Lee, ``Arbitrary style transfer with style-attentional networks,'' in \emph{proceedings of the IEEE/CVF conference on computer vision and pattern recognition}, 2019, pp. 5880--5888.

\bibitem{39li2020celeb}
Y.~Li, X.~Yang, P.~Sun, H.~Qi, and S.~Lyu, ``Celeb-df: A large-scale challenging dataset for deepfake forensics,'' in \emph{Proceedings of the IEEE/CVF conference on computer vision and pattern recognition}, 2020, pp. 3207--3216.

\bibitem{40google2019deepfakedetection}
\BIBentryALTinterwordspacing
Deepfakedetection. Accessed: Oct. 2023. [Online]. Available: \url{https://ai.googleblog.com/2019/09/contributing-data-to-deepfake-detection.html}
\BIBentrySTDinterwordspacing

\bibitem{41dolhansky2020deepfake}
B.~Dolhansky, J.~Bitton, B.~Pflaum, J.~Lu, R.~Howes, M.~Wang, and C.~C. Ferrer, ``The deepfake detection challenge (dfdc) dataset,'' \emph{arXiv preprint arXiv:2006.07397}, 2020.

\bibitem{42dolhansky2019deepfake}
B.~Dolhansky, R.~Howes, B.~Pflaum, N.~Baram, and C.~C. Ferrer, ``The deepfake detection challenge (dfdc) preview dataset,'' \emph{arXiv preprint arXiv:1910.08854}, 2019.

\bibitem{43thies2016face2face}
J.~Thies, M.~Zollhofer, M.~Stamminger, C.~Theobalt, and M.~Nie{\ss}ner, ``Face2face: Real-time face capture and reenactment of rgb videos,'' in \emph{Proceedings of the IEEE conference on computer vision and pattern recognition}, 2016, pp. 2387--2395.

\bibitem{44FaceSwap}
\BIBentryALTinterwordspacing
Faceswap. Accessed: Oct. 2023. [Online]. Available: \url{https://github.com/MarekKowalski/FaceSwap}
\BIBentrySTDinterwordspacing

\bibitem{45thies2019deferred}
J.~Thies, M.~Zollh{\"o}fer, and M.~Nie{\ss}ner, ``Deferred neural rendering: Image synthesis using neural textures,'' \emph{Acm Transactions on Graphics (TOG)}, vol.~38, no.~4, pp. 1--12, 2019.

\bibitem{46sagonas2016300}
C.~Sagonas, E.~Antonakos, G.~Tzimiropoulos, S.~Zafeiriou, and M.~Pantic, ``300 faces in-the-wild challenge: Database and results,'' \emph{Image and vision computing}, vol.~47, pp. 3--18, 2016.

\bibitem{47kingma2014adam}
D.~P. Kingma and J.~Ba, ``Adam: A method for stochastic optimization,'' \emph{arXiv preprint arXiv:1412.6980}, 2014.

\bibitem{48qian2020thinking}
Y.~Qian, G.~Yin, L.~Sheng, Z.~Chen, and J.~Shao, ``Thinking in frequency: Face forgery detection by mining frequency-aware clues,'' in \emph{European conference on computer vision}.\hskip 1em plus 0.5em minus 0.4em\relax Springer, 2020, pp. 86--103.

\bibitem{49liu2021spatial}
H.~Liu, X.~Li, W.~Zhou, Y.~Chen, Y.~He, H.~Xue, W.~Zhang, and N.~Yu, ``Spatial-phase shallow learning: rethinking face forgery detection in frequency domain,'' in \emph{Proceedings of the IEEE/CVF conference on computer vision and pattern recognition}, 2021, pp. 772--781.

\bibitem{50ni2022core}
Y.~Ni, D.~Meng, C.~Yu, C.~Quan, D.~Ren, and Y.~Zhao, ``Core: Consistent representation learning for face forgery detection,'' in \emph{Proceedings of the IEEE/CVF conference on computer vision and pattern recognition}, 2022, pp. 12--21.

\bibitem{51gu2022hierarchical}
Z.~Gu, T.~Yao, Y.~Chen, S.~Ding, and L.~Ma, ``Hierarchical contrastive inconsistency learning for deepfake video detection,'' in \emph{European Conference on Computer Vision}.\hskip 1em plus 0.5em minus 0.4em\relax Springer, 2022, pp. 596--613.

\bibitem{52haliassos2022leveraging}
A.~Haliassos, R.~Mira, S.~Petridis, and M.~Pantic, ``Leveraging real talking faces via self-supervision for robust forgery detection,'' in \emph{Proceedings of the IEEE/CVF Conference on Computer Vision and Pattern Recognition}, 2022, pp. 14\,950--14\,962.

\bibitem{53wang2023dynamic}
Y.~Wang, K.~Yu, C.~Chen, X.~Hu, and S.~Peng, ``Dynamic graph learning with content-guided spatial-frequency relation reasoning for deepfake detection,'' in \emph{Proceedings of the IEEE/CVF Conference on Computer Vision and Pattern Recognition}, 2023, pp. 7278--7287.

\bibitem{54tan2023deepfake}
L.~Tan, Y.~Wang, J.~Wang, L.~Yang, X.~Chen, and Y.~Guo, ``Deepfake video detection via facial action dependencies estimation,'' in \emph{Proceedings of the AAAI Conference on Artificial Intelligence}, vol.~37, no.~4, 2023, pp. 5276--5284.

\bibitem{55van2008visualizing}
L.~Van~der Maaten and G.~Hinton, ``Visualizing data using t-sne.'' \emph{Journal of machine learning research}, vol.~9, no.~11, 2008.

\end{thebibliography}

\vspace{11pt}

\end{sloppypar}
\end{document}